\documentclass[10pt,twocolumn,letterpaper]{article}

\usepackage{cvpr}              

\usepackage{multirow}

\usepackage{graphicx}
\usepackage{amsmath}
\usepackage{amssymb}
\usepackage{booktabs}
\usepackage{mathabx}
\usepackage[utf8]{inputenc} 
\usepackage[title]{appendix}

\usepackage[pagebackref,breaklinks,colorlinks]{hyperref}

\usepackage[capitalize]{cleveref}
\crefname{section}{Sec.}{Secs.}
\Crefname{section}{Section}{Sections}
\Crefname{table}{Table}{Tables}
\crefname{table}{Tab.}{Tabs.}

\def\cvprPaperID{8} 
\def\confName{CVPR}
\def\confYear{2022}

\begin{document}

\title{DooDLeNet: Double DeepLab Enhanced Feature Fusion for Thermal-color Semantic Segmentation}

\author{Oriel Frigo\\
\and
Lucien Martin-Gaffé\\ \\
AnotherBrain\\
Paris, France\\
{\tt\small \{oriel, lucien, catherine\}@anotherbrain.ai}
\and
Catherine Wacongne\\
}

\maketitle

\begin{abstract}
In this paper we present a new approach for feature fusion between RGB and LWIR Thermal images for the task of semantic segmentation for driving perception. We propose DooDLeNet, a double DeepLab architecture with specialized encoder-decoders for thermal and color modalities and a shared decoder for final segmentation. We combine two strategies for feature fusion: confidence weighting and correlation weighting. We report state-of-the-art mean IoU results on the MF dataset \cite{Ha2017}.
\end{abstract}

\section{Introduction}
\label{sec:intro}

Thermal images can improve the performance of computer vision algorithms, particularly during night time or in harsh weather conditions, when visibility becomes limited.
On the other hand, color images provide a richer source of visual information during day time. Hence it can be highly beneficial for vision models to combine both color and thermal images into a RGBT (red, green, blue, thermal) image pipeline.

Multimodality or the combination of different imaging sources becomes crucial to alleviate perception and ensure robustness under poor visibility conditions, but the combination of different cameras and sensors comes with alignment and synchronization challenges. Calibration of thermal images is a painstaking task where special heated or masked calibration patterns need to be employed.
Alignment of color and thermal images is also a challenging problem, not only because of common differences in field of view and resolution  between the color and thermal cameras but also because of the huge differences in image appearance between the two imaging modalities. 

In this paper we consider the task of semantic segmentation of paired color RGB and thermal LWIR (Long Wave Infra-Red) images. Most previous research in this field has been focused in searching for good fusion strategies between the color and thermal images. However, previous work on semantic color-thermal segmentation has barely considered how to fuse the images taking into account the feature disagreement due to poor alignment.

Inspired by recent advances in semantic matching, here we consider an end-to-end deep learning strategy which indirectly takes into account the alignment of RGB and LWIR thermal images through a matching correlation layer.

We show the benefits of the correlation and confidence aware feature fusion for the task of semantic segmentation on RGB-thermal data. 

The main contributions of this paper are the following:

\begin{itemize}
    \item We propose a new CNN model for color-thermal semantic segmentation. We show the interest of learning deep features that are specialized for both thermal images and color images with specialized encoder-decoder paths. 
    \item We show that semantic segmentation performance on color-thermal image pairs is increased by the aid of two different weighted feature fusion strategies: 
    \begin{itemize}
        \item For matching correlation weighting, we explicitly estimate the feature matching at the last specialized decoder layer by computing feature correlations.
        \item For confidence weighting, we estimate the segmentation confidence at both color and thermal features, and this confidence is used to weight differently the color and thermal features. 
    \end{itemize}   
\end{itemize}

\section{Literature Review}

\subsection{Semantic segmentation}

Autonomous driving requires understanding the scene ahead. The release of datasets BDD100K \cite{DBLP:conf/cvpr/YuCWXCLMD20} and Cityscapes \cite{Cordts2016Cityscapes}  has boosted the interest of using semantic segmentation as a way to process images for driving applications. Semantic segmentation consists in assigning a class to every pixel of an image. Early efforts in deep-learning based semantic segmentation used fully convolutional network (FCN) \cite{DBLP:journals/pami/ShelhamerLD17} which allowed for dense pixel classification of arbitrary sized images. However, FCN generated pixel labels from high level feature maps, resulting in relatively imprecise object boundaries. Subsequently, one of the main challenge of this task was to propose methods that extracted global context while providing pixel precise information to recover both the semantic categories present in the image and the pixel level boundaries of each object. Several approaches were developed to do so. 

Modifying the original FCN design, encoder/decoder architectures were proposed, that compress information to capture the global context before decompressing it using either deconvolution \cite{noh2015learning}, pooling indexes (SegNet)\cite{journals/corr/BadrinarayananK15} or skip connections (U-net) \cite{ronneberger2015u} to recover the fine spatial information. 

Other methods use pyramids of features \cite{farabet2012learning, zhao2017pyramid} extracted by an encoder model applied to the same image at different resolutions. The features maps can then be upsampled to match full scale the image size. 
More recent work proposed to use atrous convolutions \cite{Chen_2018_ECCV, chen2017deeplab} to capture features at different spatial scales without the need to upsample the feature maps. Combining atrous convolutions and spatial pyramid pooling, Deeplabv3+ \cite{Chen_2018_ECCV} proposes a simple and effective encoder-decoder FCN architecture. 

Another approach consists in introducing attention mechanisms \cite{fu2019dual, yu2020context, xie2021segformer} to combine different levels of feature maps and better capture long-range image dependencies. 

\subsection{Multimodality and driving perception}

The release of the KAIST Multispectral dataset \cite{hwang2015multispectral} can be seen as an important step in the direction of combining color and thermal images for driving perception.
The dataset provides physically aligned color and thermal images with bounding boxes annotations for a few semantic classes representing humans, such as \emph{person}, \emph{people} and \emph{cyclist}.

Besides the release of the KAIST dataset, other works have considered the problem of multimodality in driving perception. The work of \cite{Abbott2020} proposed a cycle consistent GAN which generates synthetic color images from LWIR and synthetic LWIR from color images while enforcing that object detection remains accurate in both modalities. The authors obtain the best object detection accuracy when combining synthetic RGB and true LWIR images.

It should be noted that thermal and color images are often not perfectly aligned, due to the parallax effect, poor calibration or registration.
This implies that naive input fusion, which consists in simply stacking the color and thermal images together by forming a 4-channel RGBT data input is rarely the most effective fusion strategy. 

A number of works investigated better strategies to fuse different image modalities together in one model. Fusenet \cite{DBLP:conf/accv/HazirbasMDC16} was one of the first works dealing indirectly with the multimodal fusion problem, by proposing a double encoder architecture for the fusion of depth maps and color images. 

The work of \cite{Ha2017} proposed the MFNet (Multi-spectral Fusion Networks), a method for semantic segmentation on RGBT data. Their model consists of separate encoders for RGB and thermal data and fusion at the decoder level. The authors also published a semantic segmentation dataset on RGBT data, the MF dataset. The RTFNet method \cite{Sun2019} performs semantic segmentation on RGBT data by fusing RGB and thermal features at multiple network depths, the method was also evaluated on the MF dataset \cite{Ha2017}. Fusion of features extracted at multiple CNN layers as performed by \cite{Sun2019} can be seen a strategy to attenuate feature disagreements, as the misalignment magnitude is decreased in deeper network layers.

FuseSeg-161 \cite{FuseSeg}, a multi-modal fusion model has been able to fuse feature from thermal and RGB images in one end-to-end model. It consists of two encoders to extract features from input, one for RGB and another for thermal images with a two-stage fusion (TSF) strategy. In the first stage, the thermal feature maps are hierarchically added with the RGB feature maps in the RGB encoder. The fused feature maps are then concatenated with the corresponding decoder feature maps in the second fusion stage. TSF method achieved over state-the-art semantic segmentation performance under various lighting conditions.

More recently, FEANet \cite{DBLP:journals/corr/abs-2110-08988} proposed a new way to extract detail spatial information with a two-stage Feature-Enhanced Attention Network. The Feature-Enhanced Attention Module (FEAM) improve excavation feature capacity from both the channel and spatial views, involving a better preservation of spatial information. Thus, the model shift more attention to high-resolution features which outperform other state-of-the-art techniques.

Another approach consists in training a deep neural network that weights differently thermal and color features. A first work in this direction was done by \cite{DBLP:conf/iccv/ZhangZC0LL19}, where the model learns to estimate a position shift between pedestrian bounding boxes in color and thermal modalities.

Our work is related to the method proposed by \cite{DBLP:conf/iccv/ZhangZC0LL19} in the sense of performing a sort of confidence feature weighting. However, while \cite{DBLP:conf/iccv/ZhangZC0LL19} is focused on bounding box retrieval and weighting, we compute dense confidence and correlation weights that are better suited for semantic segmentation.

\section{Proposed method}

\begin{figure*}[t]
\includegraphics[width=14cm]{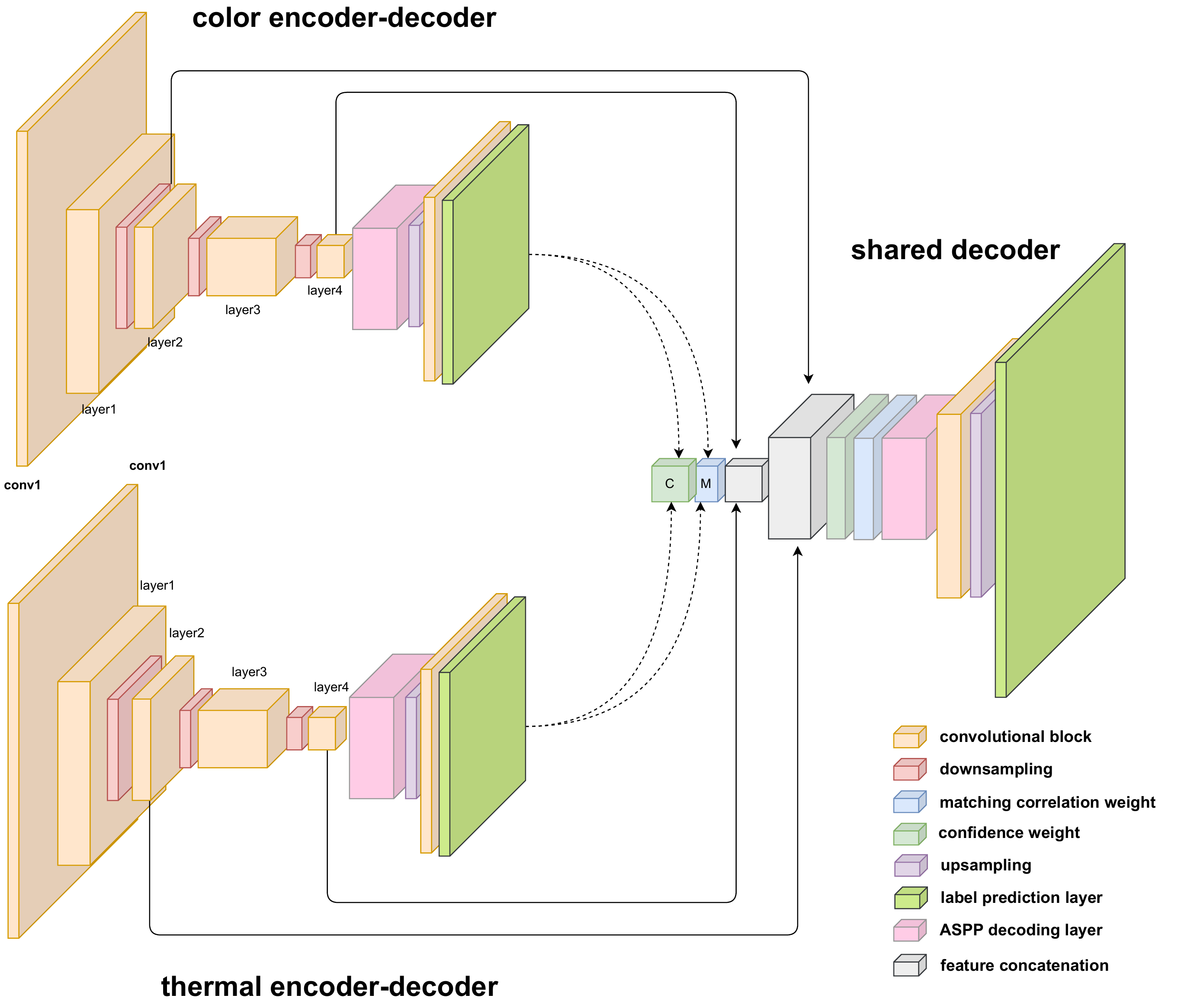}
\centering
\caption{Summarized illustration of model architecture for weighted fusion semantic segmentation. The RGB and thermal encoder-decoder paths are specialized and both provide a rough segmentation used to compute a spatial confidence score. The confidence and matching correlation is used to weight the RGB and thermal features before fusing them in the final shared decoder. Solid arrows represent concatenation-based skip connections, dashed arrows represent data flow. 
Weighted concatenated fusion is performed with features coming from two different encoder depths (layer 2 and layer 4). }
\label{fig:model}
\end{figure*}

Given a dataset of paired color and thermal images with pixel-level target labels, we seek to perform semantic segmentation on this data by fusing color and thermal images in a manner that maximizes overall segmentation performance. Here we propose a mid-level feature fusion approach, which aims to limit the feature disagreements caused by misalignment between the thermal and color images.

Differently than the work of \cite{Sun2019} which performs fusion of RGB and thermal features directly at the RGB encoder, we employ completely specialized thermal and color encoders.

We rely upon the DeepLabV3+ \cite{Chen_2018_ECCV} model backbone, which has dilated (atrous) convolutions, limiting the loss of spatial resolution and better preserving structural details. The main decoder ingredient is the ASPP (Atrous Spatial Pyramid Pooling) which obtains multi-scale features obtained by atrous convolutions with different dilation rates. 

We propose a double encoder-decoder variant of DeepLabV3+ where fusion is performed at two different feature resolutions at the decoder side, with a shared decoder for the final segmentation. A motivation for our approach comes from the observation that rich encoder features from pretrained networks such as ResNet \cite{DBLP:conf/cvpr/HeZRS16} could be adapted as seamless as possible for multiple deep learning tasks, in order to minimize training time and maximize prediction performance. 

Moreover, we perform a weighted feature fusion which takes into account feature agreement and feature confidence. Our proposed architecture is illustrated in Fig. \ref{fig:model}. We detail the two weighting strategies in the following subsections.

\subsection{Confidence weighted fusion}

It can be claimed that direct fusion of RGB and thermal features may be problematic depending on the time of day, weather and visibility conditions of the captured scene. 

For instance, during day time and favorable weather, RGB features could provide the best features to perform segmentation, as thermal images are only one-channel and texture details from objects are hardly noticeable.

On the other hand, during night time and at low light, thermal images may provide better features to segment hardly visible objects, in particular the ones with distinguishable changes in temperature compared to the background such as pedestrians and cars.

If we always give equal importance for features coming from both the modalities, the model may achieve lower performance in segmenting objects when one of the modalities provide unreliable features.

To alleviate this issue, we propose a confidence-based weighting before performing feature fusion, such that RGB and the thermal features receive difference importance based on their reliability. The decoder and the segmentation heads of each modality are able to perform a preliminary segmentation, from which a confidence can be estimated.

Based on the model confidence of the preliminary segmentation, we may be able to give more weight to the more reliable features.

\begin{figure}[t]
\includegraphics[width=5cm]{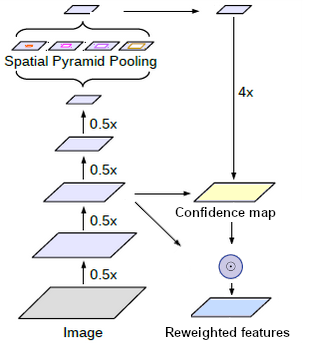}
\centering
\caption{Encoder-decoder path for modality confidence-aware feature weighting based on DeepLabv3+ architecture. The illustrated procedure is computed for both color and thermal images. Resulting re-weighted features from color and thermal are concatenated together as input for the shared decoder.}
\label{fig:confidence}
\end{figure}

Figure \ref{fig:confidence} illustrates how confidence is computed for each imaging modality.

Formally, we denote the color image as $I_c$ and thermal image as $I_t$. Each image modality $I_m$, with $m \in \{c,t\}$  has a set of features computed by a specialized encoder $\mathbf{f}_m = e_m(I_m)$ and a modality prediction computed by specialized decoder $\mathbf{y}_m = d_m(\mathbf{f}_m)$.

The confidence weight matrix $C_m$ for modality $m$ is given by an estimation of the confidence on the most probable prediction:

\begin{equation}
C_{m_i} = \text{max} \left(\frac{\exp(\mathbf{y}_{m_i})}{\sum_j \exp(\mathbf{y}_{m_j})} \right)
\end{equation}
where $\mathbf{y}_{m_i}$ denotes the one-hot logits for modality $m$ at spatial position $i$.

The confidence map $C_m$ is used to re-weight the features.
Color and thermal features receive different weights, based on the confidence of the specialized color and thermal decoders. 
Weighting is performed by point-wise Hadamard product, and resulting weighted color and thermal features are concatenated together:
\begin{align}
\mathbf{f}'_{c} = C_{c} \odot \mathbf{f}_{c}
\\
\mathbf{f}'_{t} = C_{t} \odot \mathbf{f}_{t} 
\\
\mathbf{f}'_{ct} = \mathbf{f}'_c \oplus \mathbf{f}'_t .
\label{eq:confweight}
\end{align}
where $\odot$ denotes the Hadamard product and $\oplus$ denotes concatenation. 

\subsection{Matching correlation weighted fusion}

\begin{figure}[t]
  \includegraphics[width=4cm]{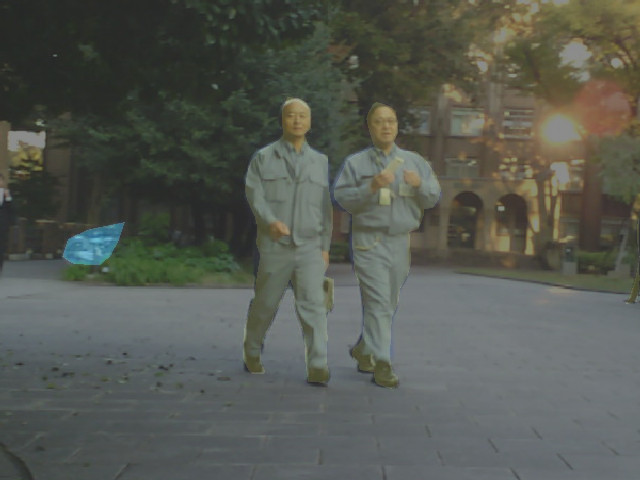}
  \includegraphics[width=4cm]{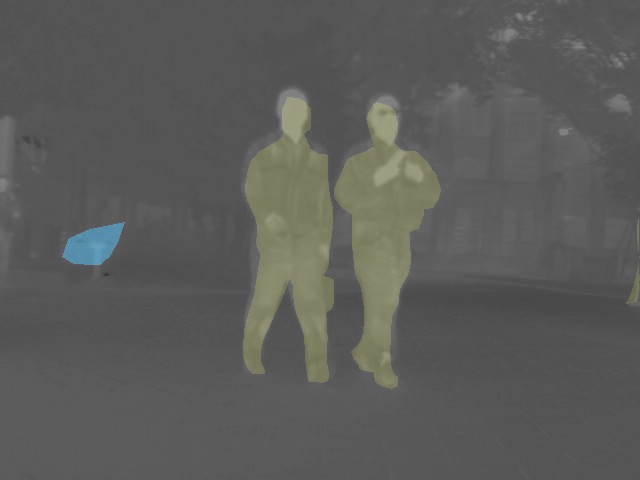}
\centering
\caption{Example of misalignment between color and thermal images. It can be seen that the label for the pedestrian class (overlayed in yellow) is not perfectly aligned with the pedestrians on the thermal image. Such misalignment can cause the RGB and the thermal encoder features to disagree. We propose a correlation feature weighting to alleviate this issue.}
\label{fig:disaligned}

\end{figure}

Another possible issue of directly fusing color and thermal features is the spatial misalignment between both images. Figure \ref{fig:disaligned} illustrates a sample image from MF dataset \cite{Ha2017} where feature disagreement can be caused due to image misalignment.

As it may be hard to achieve perfect pixel by pixel alignment between the different imaging modalities, we propose to evaluate the alignment through a matching layer.

Inspired by the work of \cite{DBLP:journals/pami/RoccoAS19} on deep image alignment,
we propose to employ a correlation layer which accounts to evaluate the matching between features coming from the two different encoders.

The correlation layer computes pair-wise feature similarities between the color and thermal decoder and the obtained scores are normalized such that weak matched features are suppressed by down weighting.

In order to deal with strong appearance differences between color and thermal features, we estimate semantic matching between the last specialized decoder, such that correlation is invariant to appearance changes.

Formally, given $\mathbf{y}_c = d_c(\mathbf{f}_c)$ corresponding to the color decoder predictions and $\mathbf{y}_t = d_t(\mathbf{f}_t)$ corresponding to the thermal decoder predictions, the correlation weight matrix $M_{ct} \in \mathbb{R}^{h \times w}$ is obtained by computing modality-wise prediction correlations (normalized scalar product of a pair of predictions), followed by a channel compressing convolutional block:

\begin{equation}
\begin{aligned}
M_{ct} = c(\| \sigma( \widebar{\mathbf{y}_t}^T \widebar{\mathbf{y}_c} )\|_2)
\end{aligned}
\label{eq:correlation}
\end{equation}

\noindent where $\sigma$ denotes ReLU activation, $\widebar{.}$ denotes spatially flattening ($y_m \in \mathbb{R}^{k \times h \times w}$, $\widebar{y_m} \in \mathbb{R}^{k \times N} , N = h \times w$, $k$ being the number of predicted classes) , $\|.\|_2$ denotes $L_2$ vector normalization and $c$ is a convolutional block performing channel compression ($N \times h\times w$ input dimensions, $ h\times w$ output dimensions). Note that $c$ acts in analogy to a feature regressor estimating parametric alignment \cite{DBLP:journals/pami/RoccoAS19}, but here we instead estimate alignment-aware spatial weights. 

Then, the concatenated confidence-weighted features obtained by Eq. \ref{eq:confweight} are re-weighted by the correlation map:
\begin{equation}
\mathbf{f}''_{ct} = M_{ct} \odot \mathbf{f}'_{ct} .
\end{equation}
Finally, the obtained feature map $\mathbf{f}''_{ct}$ is used as input to the shared decoder head to produce the final semantic segmentation.

\section{Experiments}

In this section we perform semantic segmentation experiments that shows the advantages of the proposed method.

\subsection{Dataset}

We perform semantic segmentation experiments on the MF dataset released by the authors of the method MFNet \cite{Ha2017}. To the best of our knowledge, this is the only public dataset with paired color and thermal images and dense annotated masks in the context of urban street scenes. The dataset has eight hand-labeled object classes (car, person, bike, curve, car stop, guardrail, color cone, bump) and one unlabelled background class. This dataset contains 1569 pairs of RGB and thermal images, in which 820 taken at daytime and 749 taken at nighttime. We follow the dataset splitting scheme proposed in \cite{Ha2017}, with the training set consisting of 784 pairs of images. The validation set consists of 392 pairs of images, and the test set contains 392 image pairs.

\subsection{Quantitative comparison}

In this section, we compare our model with previous works on color-thermal semantic segmentation: FRRN \cite{DBLP:journals/corr/PohlenHML16}, BiSeNet \cite{DBLP:journals/corr/abs-1808-00897}, DFN \cite{DBLP:journals/corr/abs-1804-09337}, HRNetV2 \cite{DBLP:journals/corr/abs-1904-04514}, MFNet \cite{Ha2017}, FuseNet \cite{DBLP:conf/accv/HazirbasMDC16}, DACNN \cite{DBLP:journals/corr/abs-1803-06791}, RTFNet(50-152) \cite{Sun2019}, FuseSeg-161 \cite{FuseSeg} and FEANet \cite{DBLP:journals/corr/abs-2110-08988}.

For comparison with previous work, we compute the mean IoU (Intersection Over Union) and the mean accuracy.  
The IoU (also called Jaccard index) is arguably the most adapted metric to evaluate semantic segmentation, it measures the overlap between predicted and ground truth mask, while the accuracy simply measures the total percentage of correct predictions.

In Table \ref{tab:benchmark}, it can be seen that our model improves mean IoU over all classes in the MF dataset. It improves IoU on three classes and notably on the person class, which is arguably one of the most important category for driving perception.

It should be noted that similarly to other datasets for driving perception, MF dataset is extremely class unbalanced. This implies that the IoU metric is a better indicator of segmentation performance than accuracy for this dataset.

\begin{table*}[htbp]
  \small
  \caption{Comparison with previous work on the MF test set for Acc (\%) and IoU (\%). We denote 3-channel RGB and 4-channel RGB-thermal data respectively as 3c and 4c. Note that mAcc and mIoU are calculated as a mean over all classes, including unlabelled. Bold font highlights the highest score in each column, underlined font highlights the second highest score. We report an improvement of 2\% in overall mIoU compared to previous work.}
  \setlength{\tabcolsep}{1.35mm}
    \begin{tabular}{cccccccccccccccccccccccc}
    \toprule
       \multicolumn{4}{c}{\multirow{2}*{Methods}}   & \multicolumn{2}{c}{Car } & \multicolumn{2}{c}{Person } & \multicolumn{2}{c}{Bike } & \multicolumn{2}{c}{Curve } & \multicolumn{2}{c}{Carstop } & \multicolumn{2}{c}{Guardrail } & \multicolumn{2}{c}{Cone } & \multicolumn{2}{c}{Bump} & \multirow{2}*{mAcc}  & \multirow{2}*{mIoU} \\
       \cmidrule{5-20}    \multicolumn{4}{c}{}          & Acc   &  IoU  & Acc   &  IoU  & Acc   &  IoU  & Acc   &  IoU  & Acc   &  IoU  & Acc   &  IoU  & Acc   &  IoU  & Acc   &  IoU  &  \\
    \midrule
    \multicolumn{4}{c}{FRRN-4c \cite{DBLP:journals/corr/PohlenHML16}}  & 81.9  & 74.7  & 66.2  & 60.8  & 62.8  & 50.3  & 41.2  & 35.0  & 12.5  & 11.5  & 0.0   & 0.0   & 37.2  & 34.0  & 35.2  & 34.6  & 48.5  & 44.2  \\
    \multicolumn{4}{c}{FRRN-3c \cite{DBLP:journals/corr/PohlenHML16}}  & 80.0  & 71.2  & 53.0  & 46.1  & 65.1  & 53.0  & 34.0  & 27.1  & 21.6  & 19.1  & 0.0   & 0.0   & 34.7  & 32.5  & 36.2  & 30.5  & 47.1  & 41.8  \\
    \midrule
    \multicolumn{4}{c}{BiSeNet-4c \cite{DBLP:journals/corr/abs-1808-00897}} & 89.7  & 84.1  & 72.0  & 63.2  & 74.1  & 60.1  & 45.1  & 36.7  & 34.2  & 25.3  & 18.2  & 5.0   & 47.4  & 42.2  & 39.8  & 35.9  & 57.7  & 50.0  \\
    \multicolumn{4}{c}{BiSeNet-3c \cite{DBLP:journals/corr/abs-1808-00897}} & 90.0  & 84.5  & 65.0  & 54.3  & 75.0  & 61.4  & 32.1  & 25.7  & 32.3  & 26.2  & 3.2   & 0.9   & 49.6  & 43.3  & 48.1  & 40.5  & 54.9  & 48.2  \\
    \midrule
    \multicolumn{4}{c}{DFN-4c \cite{DBLP:journals/corr/abs-1804-09337}}   & 90.0  & 84.4  & 73.2  & 65.0  & 75.5  & 60.9  & 54.0  & 40.4  & \textbf{38.9}  & 25.7  & 10.2  & 2.7   & 48.3  & 42.5  & 55.8  & 47.4  & 60.5  & 52.0  \\
    \multicolumn{4}{c}{DFN-3c \cite{DBLP:journals/corr/abs-1804-09337}}   & 90.7  & 81.4  & 67.7  & 52.8  & 71.5  & 57.5  & 49.2  & 34.9  & 35.1  & 23.8  & 4.1   & 1.4   & 44.2  & 31.0  & 54.6  & 47.5  & 57.3  & 47.5  \\
    \midrule
    \multicolumn{4}{c}{HRNet2-4c \cite{DBLP:journals/corr/abs-1904-04514}} & 92.8  & 87.6  & 79.3  & 71.0  & \underline{78.3}  & \underline{63.4}  & 59.8  & 42.5  & 25.7  & 19.1  & 18.8  & 0.0   & 56.5  & 49.8  & 63.5  & 44.5  & 63.7  & 53.2  \\
    \multicolumn{4}{c}{HRNet2-3c \cite{DBLP:journals/corr/abs-1904-04514}} & 92.2  & 86.6  & 73.1  & 59.8  & 74.9  & 61.3  & 47.0  & 33.2  & 23.8  & 28.7  & 7.3   & 0.0   & 54.6  & 47.2  & 61.5  & 46.2  & 60.9  & 51.3  \\
    \midrule
    \multicolumn{4}{c}{MFNet \cite{Ha2017}}     & 77.2  & 65.9  & 67.0  & 58.9  & 53.9  & 42.9  & 36.2  & 29.9  & 19.1  & 9.9   & 0.1   & \textbf{8.5}   & 30.3  & 25.2  & 30.0  & 27.7  & 45.1  & 39.7  \\
    \midrule
    \multicolumn{4}{c}{FuseNet \cite{DBLP:conf/accv/HazirbasMDC16}}   & 81.0  & 75.6  & 75.2  & 66.3  & 64.5  & 51.9  & 51.0  & 37.8  & 28.7  & 15.0  & 0.0   & 0.0   & 31.1  & 21.4  & 51.9  & 45.0  & 52.4  & 45.6  \\
    \midrule
    \multicolumn{4}{c}{DACNN \cite{DBLP:journals/corr/abs-1803-06791}} & 85.2  & 77.0  & 61.7  & 53.4  & 76.0  & 56.5  & 40.2  & 30.9  & 9.9   & \underline{29.3}  & 22.8  & 6.4  & 32.9  & 30.1  & 36.5  & 32.3  & 55.1  & 46.1  \\
    \midrule
    \multicolumn{4}{c}{RTFNet-50 \cite{Sun2019}} & 91.3  & 86.3  & 78.2  & 67.8  & 71.5  & 58.2  & \textbf{69.8}  & 43.7  & 32.1  & 24.3  & 13.4  & 3.6   & 40.4  & 26.0  & 73.5  & \underline{57.2}  & 62.2  & 51.7  \\
    \midrule
    \multicolumn{4}{c}{RTFNet-152 \cite{Sun2019}} & 93.0  & 87.4  & 79.3  & 70.3  & 76.8  & 62.7  & 60.7  & 45.3  & \underline{38.5}  & \textbf{29.8} & 0.0   & 0.0   & 45.5  & 29.1  & 74.7  & 55.7  & 63.1  & 53.2  \\
    \midrule
    \multicolumn{4}{c}{FuseSeg-161 \cite{FuseSeg}} & \underline{93.1}  & \textbf{87.9}  & \underline{81.4}  & \underline{71.7} & \textbf{78.5} & \textbf{64.6} & \underline{68.4} & 44.8  & 29.1  & 22.7  & \underline{63.7}  & 6.4   & 55.8  & 46.9  & 66.4  & 47.9  & \underline{70.6}  & 54.5  \\
    \midrule
    \multicolumn{4}{c}{FEANet \cite{DBLP:journals/corr/abs-2110-08988}}     & \textbf{93.3} & \underline{87.8} & \textbf{82.7} & 71.1  & 76.7  & 61.1  & 65.5  & \underline{46.5} & 26.6  & 22.1  & \textbf{70.8} & \underline{6.6} & \textbf{66.6} & \textbf{55.3} & \textbf{77.3}  & 48.9  & \textbf{73.2} & \underline{55.3} \\
\midrule
    \multicolumn{4}{c}{Ours} & 91.7 & 86.7 & 81.3 & \textbf{72.2} & 76.6 & 62.5 & 58.9 & \textbf{46.7} & 36.2 & 28.0 & 35.2 & 5.1 & \underline{56.9} & \underline{50.7} & \underline{74.8} & \textbf{65.8} & 67.9 & \textbf{57.3}  \\
    \bottomrule
    \end{tabular}%
  \label{tab:benchmark}%
\end{table*}%

\subsection{Qualitative comparison}

\begin{figure*}[h]
    \includegraphics[width=7.15in]{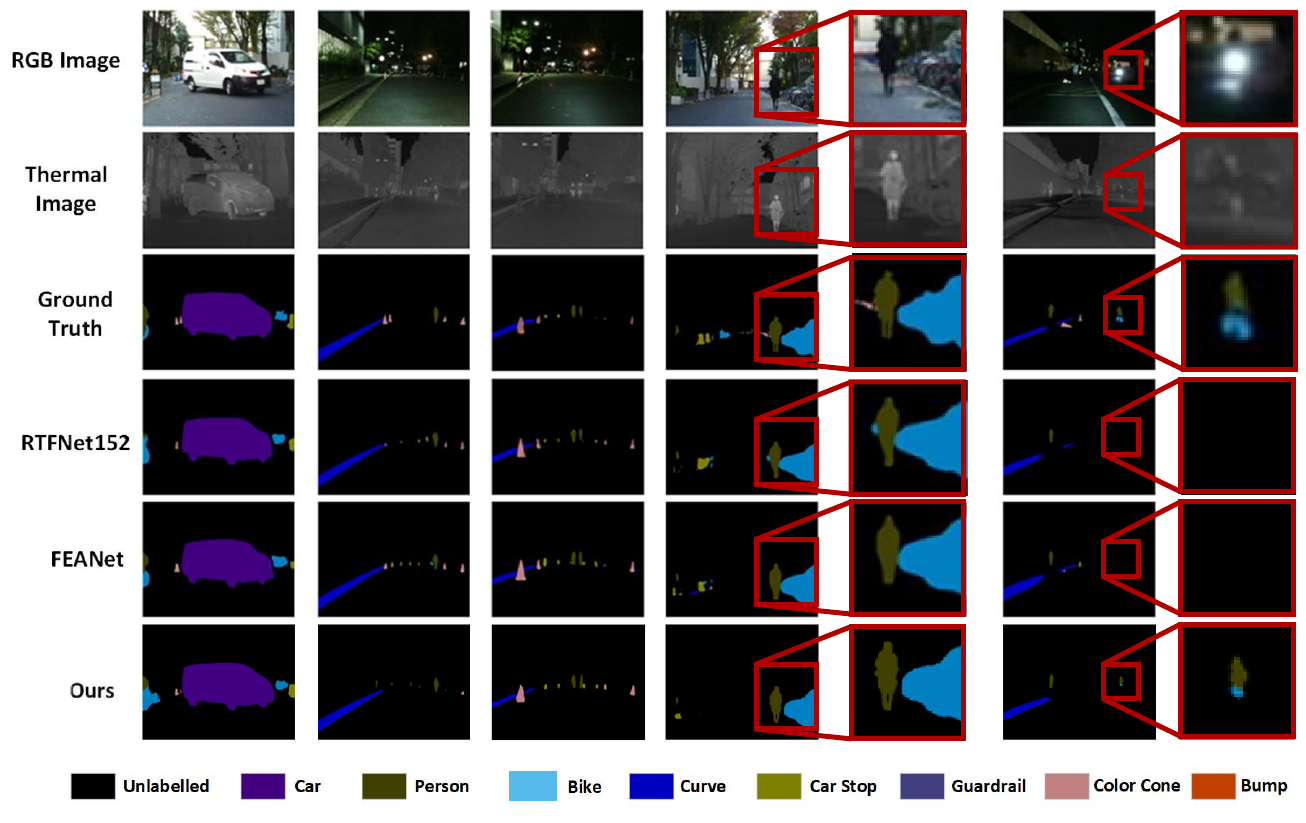}
    \vspace{-1mm}
    \caption{Qualitative results where the robustness of our method can be appreciated for daytime and nighttime scenes of MF test set split. In the last two columns (original and zoomed in images), it can be seen that our method is the only segmenting the person and the bike objects.}
    \label{fig:qualitative_zoom}
\end{figure*}

In Fig. \ref{fig:qualitative_zoom}, we show a number of semantic segmentation results obtained with our method. It can be seen that segmentation of important details are improved compared to previous work. In the zoomed in parts of columns 4 and 5, it can be seen that our method segments better the pedestrian legs.
In the last two columns, it can also be seen that our method is the only one able to retrieve the person and the bike objects, which are hardly visible in the color image due the small size of the object and a saturated highlight.

\subsection{Ablation Study}

In Table \ref{tab:ablation}, we consider a number of variants of our model, all of them also based on the DeepLabv3+ backbone. We evaluate the individual contributions of the confidence weighting, the concatenated decoder-level fusion and correlation weighting. 

We also evaluate the model performance in different lighting conditions, only day, and only night. It can be seen that our proposed model obtains the highest mIoU in all lighting conditions, compared to its simpler variants.

\subsection{Learned features visualization}

\begin{figure*}[t]
\includegraphics[width=12cm]{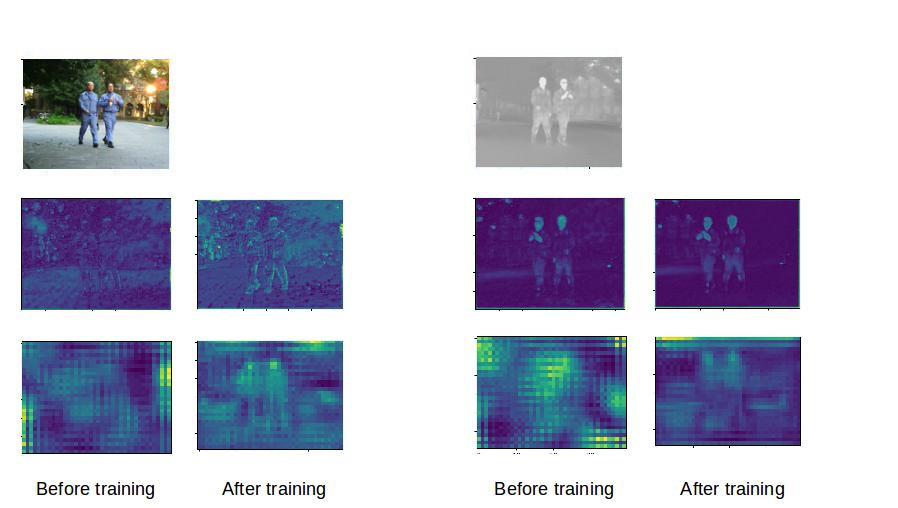}
\centering
\caption{Illustration of features learned by color encoder (left two columns) and thermal encoder (right two columns) at encoder layer 2 and layer 4 (second and third rows) before (pretrained on Imagenet) and after training for semantic segmentation. Note that at the deeper level the feature responses seem less invariant to the imaging modality after training. }
\label{fig:learned_features}
\end{figure*}

In Fig. \ref{fig:learned_features}, we visualize the mean output feature responses of our proposed model before and after semantic segmentation training on RGB-T data. It can be seen that for the given image sample, the shallow second encoder layer gives somewhat similar responses before and after training. On the other hand, the deeper feature response at deeper layer seems to vary more in its response after training, by learning to spot at the two pedestrians.  

\begin{table*}[tbp]
  \centering
  \caption{Ablation study: comparison of mAcc ($\%$) and mIoU ($\%$) on the day-only, night-only and regular test set of MF dataset for different model variants. For a fair comparison, all of the variants are trained with the same optimizer parameters and are based on DeepLabv3+ with Resnet-101 pretrained backbone. It can be noted the advantage of our approach - simpler model variants based on DeepLabv3+ result in lower mIoUs for all lighting conditions.}
  \begin{tabular*}{\hsize}{@{}@{\extracolsep{\fill}}lllllllllllll@{}}
    \midrule
    \multicolumn{3}{c}{\multirow{2}*{Variants}} & \multicolumn{3}{c}{Daytime} & \multicolumn{3}{c}{Nighttime} & \multicolumn{3}{c}{Both}\\
    \cmidrule{4-12}          &       &       & \multicolumn{1}{c}{mAcc} &       & \multicolumn{1}{c}{mIoU} & \multicolumn{1}{c}{mAcc} &       & \multicolumn{1}{c}{mIoU} & \multicolumn{1}{c}\
{mAcc} &       & \multicolumn{1}{c}{mIoU}\\
    \midrule
    \multicolumn{3}{c}{Only RGB} & 54.3  && 44.4 & 56.5 && 48.3 & 61.2 & & 50.7 \\
    \midrule
    \multicolumn{3}{c}{Only thermal} & 48.2 &      & 39.0 & 64.0 && 51.9 &  65.6 && 50.1   \\
    \midrule
    \multicolumn{3}{c}{Input fusion - stacked RGBT} & 48.7 & & 42.8 & 56.0 && 51.0 & 57.1 && 51.1  \\
    \midrule
    \multicolumn{3}{c}{Unweighted fusion} & 60.7 & & 49.0 & 61.4 && 53.6 &  64.9 && 54.7  \\
    \midrule
    \multicolumn{3}{c}{Only-confidence weighting} & \textbf{63.3} && 49.6 & 61.2 && 54.1 & 65.4 && 55.2 \\
    \midrule
    \multicolumn{3}{c}{Ours} & 58.8 &       & \textbf{50.1} & \textbf{64.1} & & \textbf{55.7}  & \textbf{67.9} & & \textbf{57.3} \\
    \bottomrule
  \end{tabular*}%
  \label{tab:ablation}%
\end{table*}%

\subsection{Implementation and setup details}

Our method is implemented using the pytorch library \cite{NEURIPS2019_9015} and the torch segmentation models library \cite{Yakubovskiy:2019}. For all experiences with our method, we use ResNet-101 model \cite{DBLP:conf/cvpr/HeZRS16} pretrained on Imagenet as backbone. We perform concatenation and confidence weighting with features from layer 2 and layer 4 of DeepLabv3+ encoder. Bilinear interpolation is used to downsample or upsample confidence map $C_m$ and matching map $M_{ct}$ whenever needed to fit the encoder feature spatial resolutions from layer 2 and layer 4.

For a fair comparison, we perform similar training setup as compared to previous methods. We adopt mostly the same hyperparameters as employed by the authors of the RTFNet method \cite{Sun2019}. We use random cropping and random flipping as image augmentation during training.

For optimization, we use SGD with momentum set to 0.9, weight decay set to 0.0005, initial learning rate set to 0.01 and a maximum of 50 training epochs. We employ exponential decaying learning rate scheduler with gamma set to 0.95. As commonly done in the field, we minimize the cross entropy loss between the predicted and the target labels.

\section{Conclusion}

In this paper we have presented a new feature fusion approach for the task of semantic segmentation of paired color RGB and thermal LWIR images. We proposed DooDLeNet, a double DeepLab net which fuses multimodal images taking into account the feature disagreements and feature confidences. We have shown that this approach improves semantic segmentation performance compared to previous works, in particular for the correct segmentation of pedestrians. Future work may be focused on evaluating different model backbones and to refine further segmentation details.

\section{Acknowledgements}

This research is part of Heliaus project\footnote{European Union's Heliaus  project website: \url{https://heliaus.eu/}}, which has received funding from the ECSEL Joint Undertaking (JU) under grant agreement No 826131. The JU receives support from the European Union’s Horizon 2020 research and innovation programme and France, Germany, Ireland, Italy\footnote{Disclaimer: Funded by the European Union. Views and opinions expressed are however those of the authors only and do not necessarily reflect those of the European Union or ECSEL JU. Neither the European Union nor the granting authority can be held responsible for them.}. Completed by local funding from the French region Auvergne Rhône Alpes.

\bibliography{heliaus_semantic_segmentation} 
\bibliographystyle{ieeetr}

\end{document}